%% file: author.tex
\documentclass{svproc}
\usepackage{url}
\usepackage{amsmath, amssymb} 
\usepackage{graphicx} 
\usepackage{multirow}
\usepackage{enumitem}

\begin{document}
\mainmatter              
\title{High-Dimensional Bayesian Optimization with Constraints: Application to Powder Weighing}
\titlerunning{Application to Powder Weighing}  
%
\author{Shoki Miyagawa, Atsuyoshi Yano, Naoko Sawada, Isamu Ogawa}
\authorrunning{S. Miyagawa, A. Yano, N. Sawada, I. Ogawa} 
%
\tocauthor{S. Miyagawa, A. Yano, N. Sawada, I. Ogawa}
\institute{Information Technology R\&D Center, MITSUBISHI Electric Corporation, Kamakura 247-8501, Japan,\\
\email{Miyagawa.Shoki@ds.MitsubishiElectric.co.jp}
\email{Yano.Atsuyoshi@dn.MitsubishiElectric.co.jp}
\email{Sawada.Naoko@df.MitsubishiElectric.co.jp}
\email{Ogawa.Isamu@ah.MitsubishiElectric.co.jp}
}

\newcommand{\argmin}{\mathop{\rm arg~min}\limits}
\newcommand{\argmax}{\mathop{\rm arg~max}\limits}

\maketitle              

\begin{abstract} 
Bayesian optimization works effectively optimizing parameters in black-box problems.
However, this method did not work for high-dimensional parameters in limited trials.
Parameters can be efficiently explored by nonlinearly embedding them into a low-dimensional space; however, the constraints cannot be considered.
We proposed combining parameter decomposition by introducing disentangled representation learning into nonlinear embedding to consider both known equality and unknown inequality constraints in high-dimensional Bayesian optimization.
We applied the proposed method to a powder weighing task as a usage scenario.
Based on the experimental results, the proposed method considers the constraints and contributes to reducing the number of trials by approximately 66\% compared to manual parameter tuning.

\end{abstract}

\input{figure}

\input{table}

\input{1-introduction}
\input{2-relatedwork}

\input{3-method}
\input{4-dataset}
\input{5-implementation}
\input{6-evaluation}
\input{7-discussion}
\input{8-conclusion}

\noindent
\textbf{\ackname}\
The authors are grateful to Hirofumi Nitta, Masato Nakagome and Kenji Takizawa from Tsukishima Machine Sales Co Ltd and Tomoki Yoshimura from Tsukishima Kikai Co Ltd for providing the dataset and for their kind cooperation in the experiments.

\bibliographystyle{./styles/bibtex/spmpsci}
\bibliography{reference} 

\end{document}

%% file: figure.tex
\newcommand{\figframework}
{
\begin{figure}[t]
    \centering
    \includegraphics[width=\textwidth]{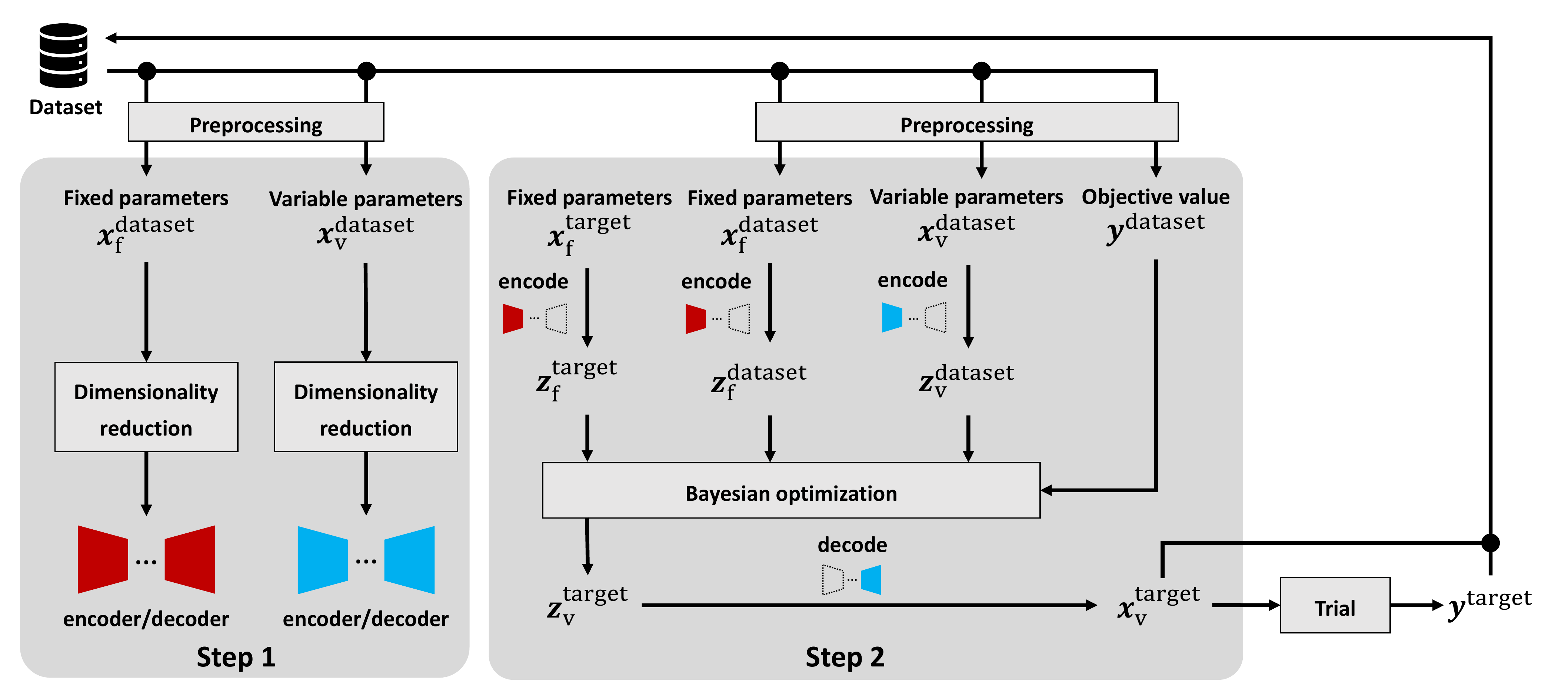}
    \caption{Overview of the proposed method.}
    \label{fig:framework}
\end{figure}
}

\newcommand{\figpowder}
{
\begin{figure}[t]
    \centering
    \includegraphics[width=0.95\textwidth]{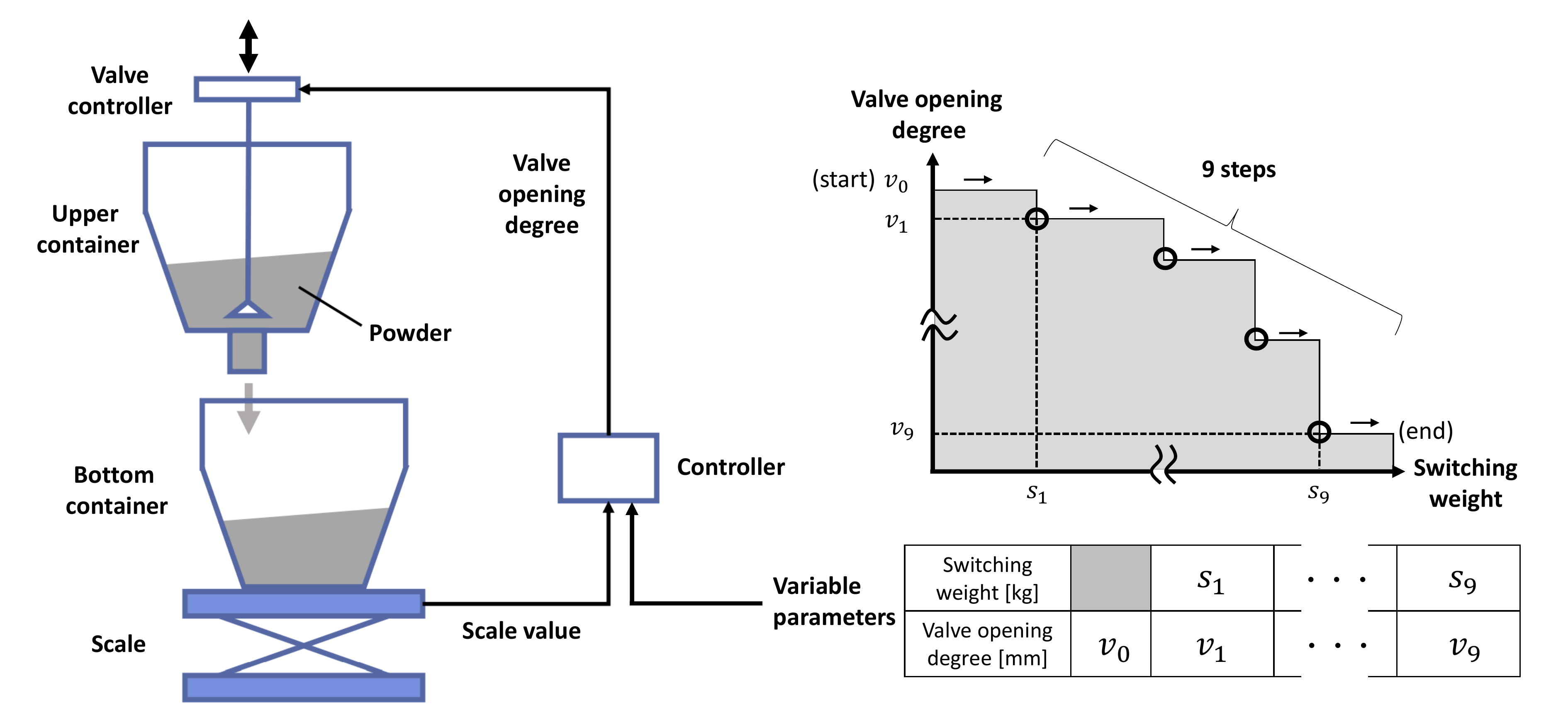}
    \caption{Powder weighing system (left) and variable parameters (right).}
    \label{fig:powder_weighing}
\end{figure}
}

\newcommand{\figexratio}
{
\begin{figure}[t]
    \centering
    \includegraphics[width=\textwidth]{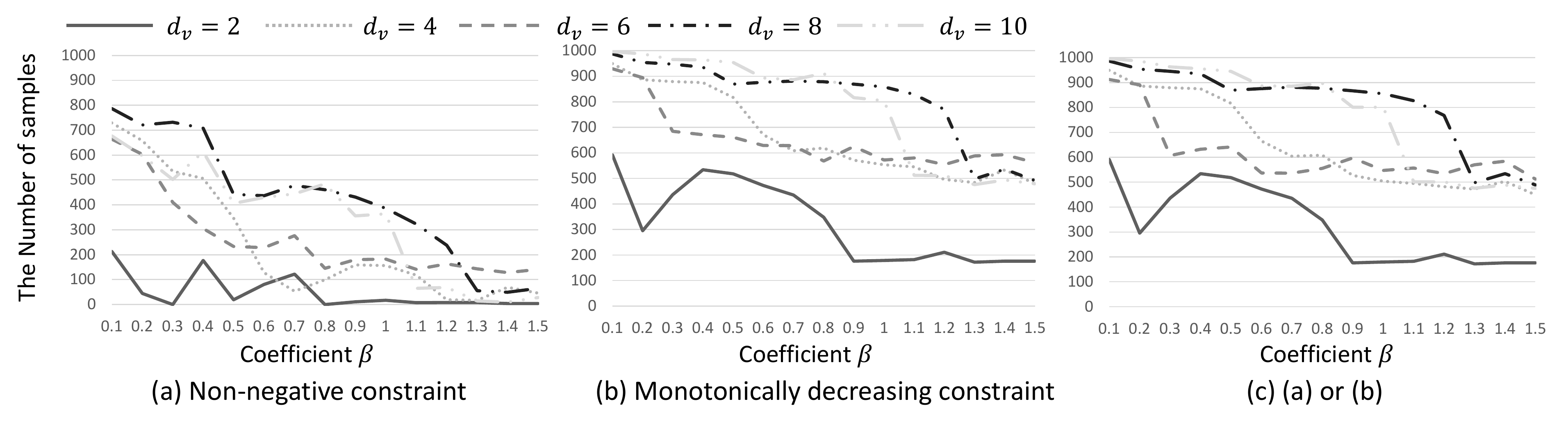}
    \caption{The number of samples that did not satisfy the following constraints. (a) non-negative constraints, (b) monotonically decreasing constraints, (c) (a) or (b).}
    \label{fig:ex1_ratio}
\end{figure}
}

\newcommand{\figexparameter}
{
\begin{figure}[t]
    \centering
    \includegraphics[width=0.95\textwidth]{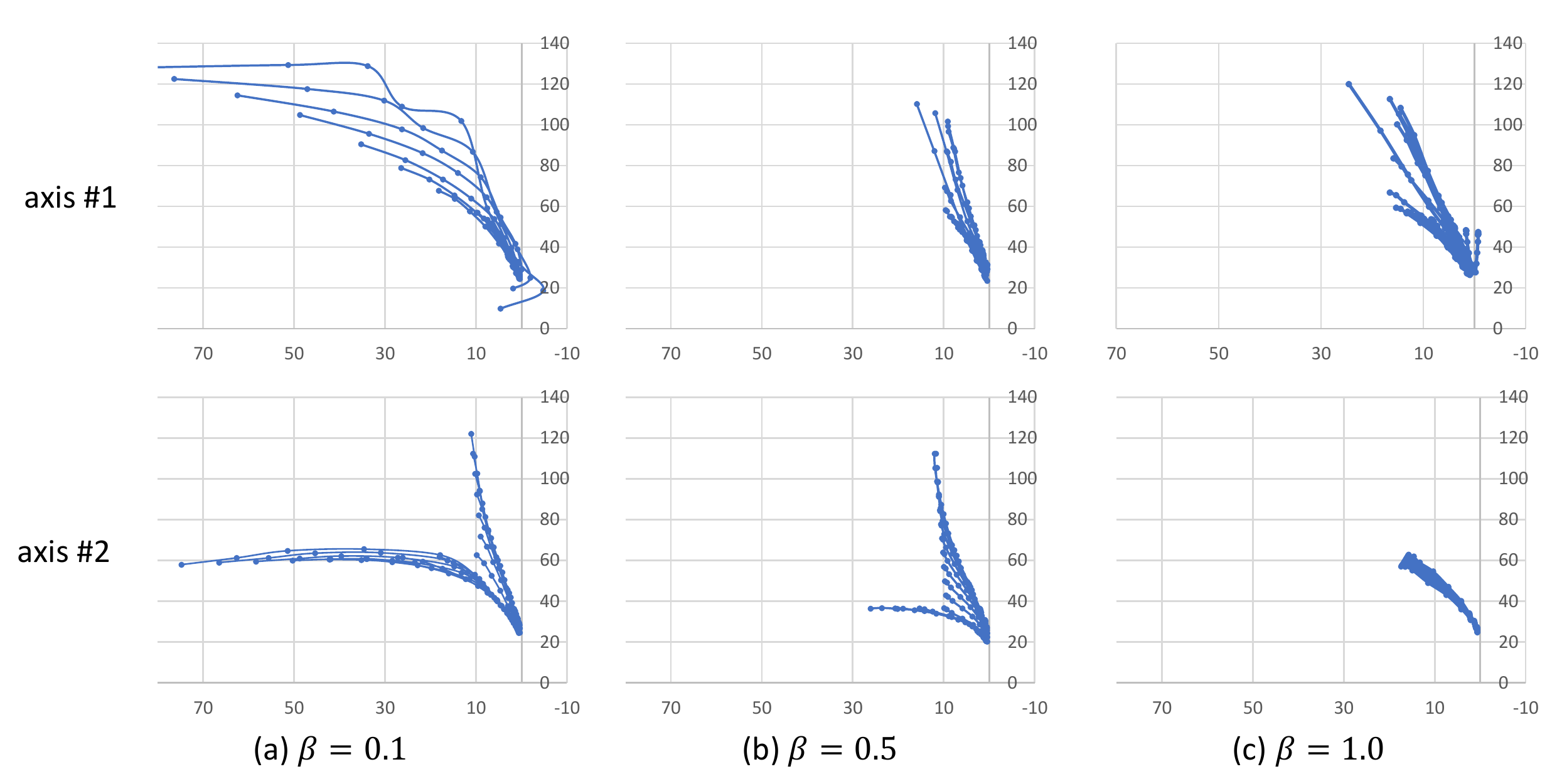}
    \caption{Comparison of variable parameters for different $\beta$. The horizontal axis represents the switching weights, whereas the vertical axis represents the valve opening degrees.}
    \label{fig:ex1_parameter}
\end{figure}
}

\newcommand{\figexsecond}
{
\begin{figure}[t]
  \centering
  \begin{minipage}[b]{0.3\linewidth}
    \centering
    \includegraphics[width=\textwidth]{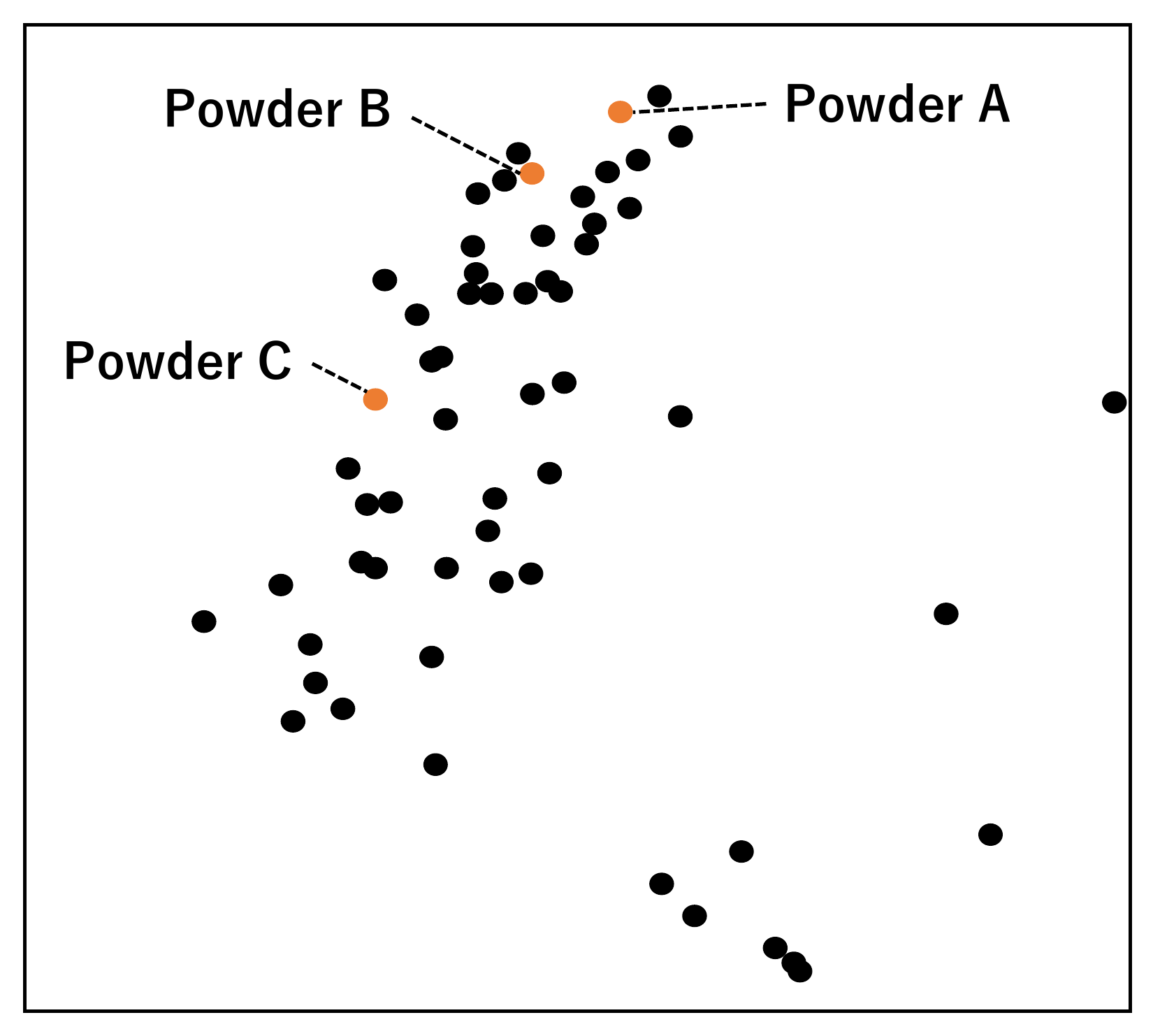}
    \caption{Mapping fixed parameters on a 2D scatterplot.}
    \label{fig:embedding}
  \end{minipage}
  \begin{minipage}[b]{0.6\linewidth}
    \centering
    \includegraphics[width=\textwidth]{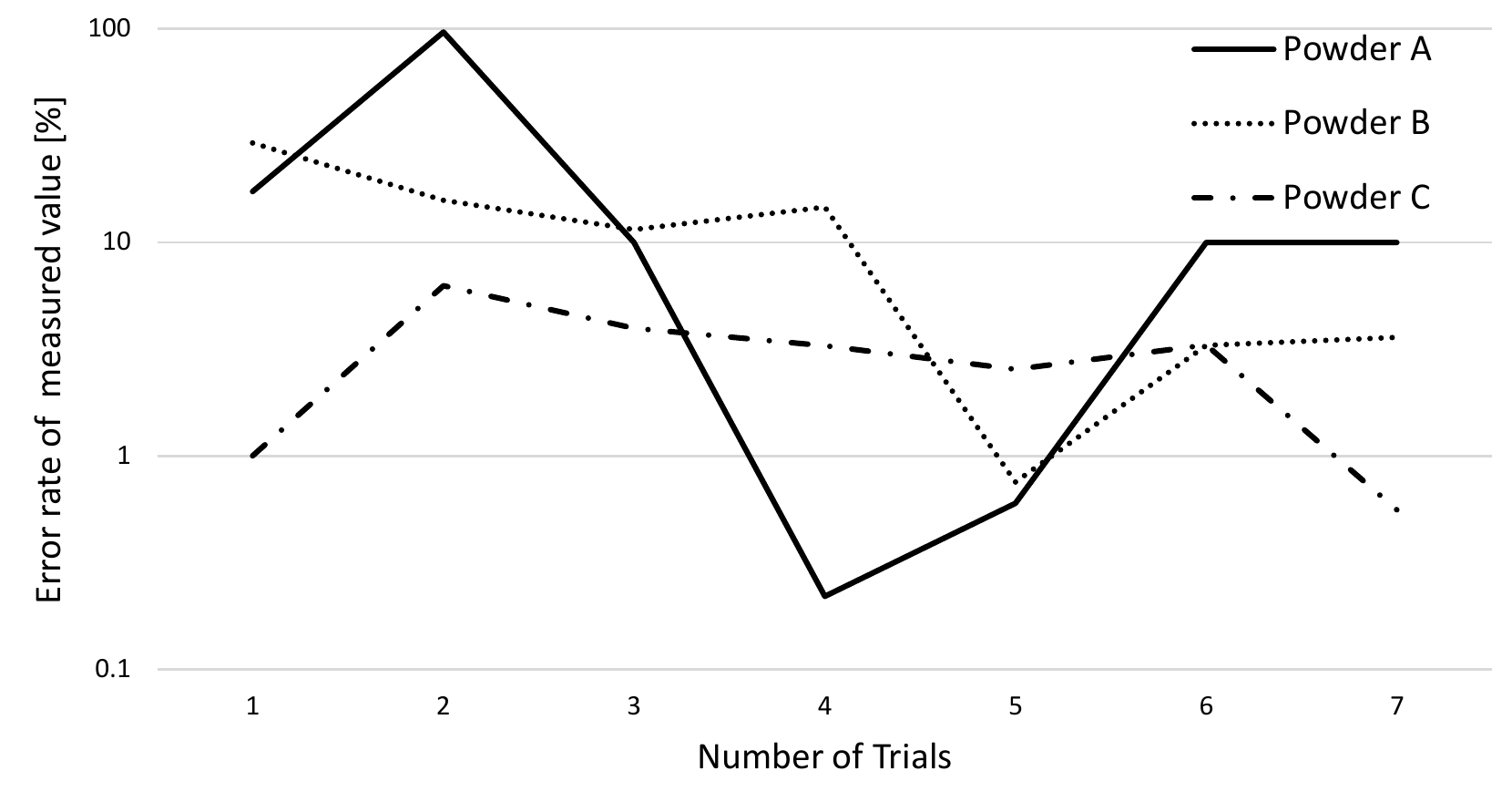}
    \caption{Weighing errors for each trial.}
    \label{fig:bo_result}
  \end{minipage}
\end{figure}
}

%% file: table.tex
\newcommand{\tabparameter}
{
\begin{table}[t]
\caption{A list of parameters in the datasets.}
\label{tab:parameter}
\begin{tabular}{cccccl}
\hline
 &
  Name &
  Type &
  Size &
  Unit &
  Definition \\ \hline
\multirow{2}{*}{\begin{tabular}[c]{@{}c@{}}Variable\\ parameters\end{tabular}} &
  Valve opening degree &
  Double &
  10 &
  mm &
  See Section 4.1 \\
 &
  Switching weight &
  Double &
  9 &
  kg &
  See Section 4.1 \\ \hline
\multirow{7}{*}{\begin{tabular}[c]{@{}c@{}}Fixed\\ parameters\end{tabular}} &
  Physical property &
  Double &
  11 &
  - &
  \begin{tabular}[c]{@{}l@{}}Parameters related \\ to the fall speed\end{tabular} \\
 &
  Required weight &
  Double &
  1 &
  kg &
  \begin{tabular}[c]{@{}l@{}}Amount of powder to \\ be weighed\end{tabular} \\
 &
  Valve diameter &
  Integer &
  1 &
  mm &
  Diameter of the valve \\
 &
  Input weight &
  Integer &
  1 &
  kg &
  \begin{tabular}[c]{@{}l@{}}Initial weight of powder \\ in the upper container\end{tabular} \\
 &
  Shaking &
  Boolean &
  1 &
  - &
  Whether to shake a valve \\
 &
  Vibration &
  Boolean &
  1 &
  - &
  \begin{tabular}[c]{@{}l@{}}Whether to vibrate the container \\ with an actuator from outside\end{tabular} \\
 &
  Pre-vibration &
  Boolean &
  1 &
  - &
  \begin{tabular}[c]{@{}l@{}}Whether to vibrate the \\ container before weighing\end{tabular} \\ \hline
\begin{tabular}[c]{@{}c@{}}Objective\\ value\end{tabular} &
  Weighing error &
  Double &
  1 &
  kg &
  \begin{tabular}[c]{@{}l@{}}Error between a measured value \\ and a required weight\end{tabular} \\ \hline
\end{tabular}
\end{table}
}

\newcommand{\tabcondition}
{
\begin{table}[t]
\centering
\caption{User-specified fixed parameters for each powder used in Experiment 2.}
\label{tab:condition}
\begin{tabular}{ccccccc}
\hline
 & Valve diameter & Required weight & Input weight & Shaking & Vibration & Pre-vibration \\ \hline
A & 150 \rm{mm} & 10 \rm{kg} & 150 \rm{kg} & False & True & False \\
B & 150 \rm{mm} & 18 \rm{kg} & 500 \rm{kg} & False & True & False \\
C & 150 \rm{mm} & 10 \rm{kg} & 200 \rm{kg} & False & True & False \\ \hline
\end{tabular}
\end{table}
}

%% file: 1-introduction.tex
\section{Introduction}\label{sec:introduction}

Bayesian optimization has attracted attention because it can estimate the optimum parameters in models where the input-output relationship is a black box.
This enables the exploration of the optimum parameters with fewer trials than with random or grid searches.
It can be applied not only to hyperparameter optimization in deep learning but to cases such as molecular design~\cite{Bombarelli18}\cite{Yan20}, robot manipulation~\cite{Letham20}\cite{Wang17}, and melody generation~\cite{Zhou21}.
Many libraries~\cite{Akiba19}\cite{Balandat20}\cite{Liaw18} can automate experiments using Bayesian optimization, whose practical applications can be seen in various domains.


However, Bayesian optimization cannot estimate optimum parameters in a high-dimensional space (typically $>$ 10) within a reasonable number of trials because of the curse of dimensionality, where the search space is explosively vast and sparse.
Many studies~\cite{Bombarelli18}\cite{Jaquier20}\cite{Kirschner19}\cite{Letham20}\cite{Lu18}\cite{Moriconi20}\cite{Pascal21}\cite{Wang13}\cite{Yan20} have focused on exploring low-dimensional space acquired by reducing dimensionality instead of directly dealing with high-dimensional space.
We refer to a low-dimensional space as a latent space.
Particularly, nonlinear dimensionality reduction~\cite{Bombarelli18}\cite{Pascal21}\cite{Yan20} realizes more efficient parameter exploration than linear dimensionality reduction~\cite{Letham20}\cite{Wang13} or subspace exploration in the original high-dimensional space~\cite{Kirschner19}.
They can also convert discrete into continuous variables to conduct gradient ascent in the latent space.
However, they cannot consider constraints for parameters because they cannot be explicitly expressed in the latent space owing to nonlinear embedding.
Although post-processing can transform the generated parameters into those that are valid for the system, dramatic transformations of the parameters decrease the exploration efficiency.
Therefore, considering these constraints is a crucial issue in high-dimensional Bayesian optimization~(HDBO).


To address this challenge, we propose a method that considers two types of constraints: known equality and unknown inequality constraints. 
We incorporate disentangled representation learning~(DRL) to consider inequality constraints even with nonlinear dimensionality reduction.
Although DRL is generally used to improve the interpretability of a generative model and enable users to control the model, we demonstrated that it can also be used to learn features from datasets that implicitly satisfy inequality constraints.
The proposed method does not require users to provide inequality constraints explicitly. 
That is, it can be applied to unstructured data, such as images and natural languages, where the explicit formulation of constraints is difficult.
Equality constraints can be considered by dividing the parameters into variable and fixed ones, as detailed in Section~\ref{sec:method}.


The proposed method can be widely applied to HDBO under known equality and unknown inequality constraints.
Although some methods consider either high-dimensionality or constraints, few studies~\cite{Bergmann20} have focused on considering both.
Typical benchmarks include a complex artificial function~\cite{Hansen20} and robot manipulation~\cite{Letham20}\cite{Wang17}; however, these constraints are not considered.
Bergmann and Graichen~\cite{Bergmann20} proposed a method to consider unknown inequality constraints; however, they did not apply their method to high-dimensional parameters but to 2D artificial test functions in their evaluation.
This study applied the proposed method to the practical task of precisely scaling a specific amount of powder with a powder weighing system as a usage scenario to optimize parameters with high dimensionality and constraints.
Previously, the powder weighing system parameters had to be manually adjusted for each type of powder, which requires expertise and time to tune.
To the best of our knowledge, this study is the first attempt to apply a machine learning technique to a powder weighing system.
In this paper, Section~\ref{sec:powderWeightingDevice} provides an overview of the powder weighing system and the dataset; Section~\ref{sec:implementation} presents the preprocessing, model architecture, and hyperparameters of the system.


Two experiments were conducted using the powder weighing system.
Based on the first experiment, the inequality constraints can be considered by adjusting the dimensionality of the latent space and regularization coefficient.
As demonstrated by the other experiment, the proposed method achieved the target performance and reduced the number of trials by 66\% compared to manual parameter tuning.
The details of each experiment are presented in Section~\ref{sec:evaluation}.

In summary, the main contributions of this study are twofold:
\begin{itemize}
    \item We designed a versatile method for HDBO under known equality and unknown inequality constraints.
    \item By introducing DRL, Bayesian optimization can implicitly consider inequality constraints without explicit user constraints; when applied to the powder-weighing system, the proposed method allows users to determine parameters that achieve the target performance within one-third of the trials compared to manual tuning.
\end{itemize}

%% file: 2-relatedwork.tex
\section{Related Work}\label{sec:related_work}

The main approaches to dealing with HDBO are twofold.
The first approach performs Bayesian optimization in a low-dimensional space embedded in high-dimensional parameters.
Wang et al.~\cite{Wang13} proposed the use of linear random matrices to embed parameters into a low-dimensional space.
Kirschner et al.~\cite{Kirschner19} performed Bayesian optimization to explore parameters on a one-dimensional subspace; the dimensions were selected from high dimensions.
Li et al.~\cite{Li17} applied a deep learning technique, called dropout, which randomly dropped out some dimensions and performed Bayesian optimization on a subspace consisting of non-dropped dimensions.
However, applying these methods to unstructured data such as images and natural languages is challenging.
For example, in the case of images, these methods do not model the correlations between neighboring pixels because the correlations are strong.
Conversely, some methods~\cite{Bombarelli18}\cite{Jaquier20}\cite{Lu18}\cite{Moriconi20}\cite{Yan20} use a neural network, which learns embedding into low-dimensional space and models correlations between input variables.
However, a neural network cannot explicitly express constraints in the transformed space.
The proposed method implicitly considers inequality constraints by introducing DRL to address the problem.


The other approach proposed by Kandasamy et al.~\cite{Kandasamy15} is based on the additive Gaussian process, where high-dimensional parameters are decomposed into low-dimensional parameter groups.
Li et al.~\cite{Li16} and Rolland et al.~\cite{Rolland18} extended the approach and loosened the assumptions regarding the decomposition.
Inspired by this approach, we decompose the parameters into two groups to consider the equality constraints, as described in Section~\ref{sec:method}, although previous methods do not consider constraints.
In summary, the proposed method is positioned between these two conventional approaches because it combines nonlinear embedding and parameter decomposition.


%% file: 3-method.tex
\section{Method}\label{sec:method}
\figframework


This section describes the manner in which the proposed method explores the optimum parameters.
Figure~\ref{fig:framework} presents an overview of the proposed method.
First, to consider equality constraints in Bayesian optimization, users need to divide the parameters into variable parameters $x_{\rm{v}}$ without equality constraints and fixed parameters $x_{\rm{f}}$ with equality constraints and objective values $y$.
The parameters to be optimized are usually considered variable parameters, 
whereas parameters for external factors, such as physical properties and temperature, are considered fixed parameters.
Although more than one objective value can exist, we consider a single variable in this study for simplicity.

The proposed method optionally performs preprocessing on each parameter in the datasets, such as normalization and outlier removal, before each step.
Step 1 trains encoders and decoders for dimensionality reduction. 
Step 2 takes user-specified fixed parameters $x_{\rm{f}}^{\rm{target}}$ and generates variable parameters $x_{\rm{v}}^{\rm{target}}$ using encoders and decoders.
Finally, users try the variable parameters $x_{\rm{v}}^{\rm{target}}$ in the next trial and then obtain new objective values $y^{\rm{target}}$.
The input-output pair $((x_{\rm{v}}^{\rm{target}}, x_{\rm{f}}^{\rm{target}}), y^{\rm{target}})$ was added to the datasets to generate variable parameters in the subsequent trials.
Users iterate this process until reasonable parameters are obtained.
The following sections detail each step shown in Figure~\ref{fig:framework}.
A specific example of the preprocessing, model architecture, and hyperparameters of the powder weighing system is described in Section~\ref{sec:implementation}.

\subsection{Step 1: Dimensionality Reduction}\label{sec:dimensionalityReduction}

The encoders and decoders for the variables and fixed parameters were trained using the datasets.
To efficiently extract more complex features and reduce dimensionality, we used a variational autoencoder~(VAE)~\cite{Kingma14} for dimensionality reduction of the variable parameters.
The use of nonlinear models for fixed parameters is needless because they are not optimized.
In the experiment described in Section~\ref{sec:evaluation}, 
we used principal component analysis~(PCA)~\cite{Jolliffe86} to reduce the dimensionality of the fixed parameters.


Introducing DRL enables us to consider the inequality constraints for variable parameters.
The DRL decomposes high-dimensional entangled features into meaningful low-dimensional features.
Therefore, users only need to check whether the inequality constraints are satisfied for the data on each axis, instead of every area of the latent space.
If the datasets contain only data that satisfy the inequality constraints, the data on each axis are likely to meet them.
Conversely, if the datasets include data that do not satisfy the inequality constraints, the data on a particular axis might not satisfy them.
Therefore, we can limit the search area on the undesirable axis in Bayesian optimization.
For simplicity, we remove trials whose variable parameters do not satisfy inequality constraints from the datasets, as detailed in Section~\ref{sec:preprocessing}.

The proposed method uses $\beta$-VAE~\cite{Higgins17} to introduce the DRL to the VAE.
Equation~(\ref{equ:betaVAE}) is the loss function $L(x,z)$ of $\beta$-VAE.
The first term is the expected value of the log-likelihood $\log p(x_{\mathrm{v}}|z_{\mathrm{v}})$ and represents a reconstruction error, whereas the second term is the Kullback-Leibler divergence $D_{\rm{KL}}$ between the approximate posterior distribution $q(z_{\mathrm{v}}|x_{\mathrm{v}})$ and the prior distribution $p(z_{\mathrm{v}})$, such as a standard normal distribution.
The larger the coefficient $\beta$ in Equation~\ref{equ:betaVAE}, the more disentangled the representation learning is accelerated; thus, inequality constraints can be considered.
Notably, a larger $\beta$ results in a more significant reconstruction error owing to the regularization effect.
Large reconstruction errors generate parameters with rough-grained features, which implies that optimum parameters might not be generated even after many trials.

\begin{equation}
    \label{equ:betaVAE}
    L(x_{\mathrm{v}},z_{\mathrm{v}}) = \mathbb{E}_{q(z_{\mathrm{v}}|x_{\mathrm{v}})}[\log p(x_{\mathrm{v}}|z_{\mathrm{v}})] - \beta D_{\mathrm{KL}}[q(z_{\mathrm{v}}|x_{\mathrm{v}})||p(z_{\mathrm{v}})]
\end{equation}

\subsection{Step 2: Bayesian Optimization}\label{sec:step2}

Step 2 generates variable parameters $x_{\rm{v}}^{\rm{target}}$ for the next trial, according to the user-specified fixed parameters $x_{\rm{f}}^{\rm{target}}$.
First, the variable and fixed parameters, $x_{\rm{v}}^{\rm{dataset}}$ and $x_{\rm{f}}^{\rm{dataset}}$, are encoded to the latent parameters, $z_{\rm{v}}^{\rm{dataset}}$ and $z_{\rm{f}}^{\rm{dataset}}$, respectively.
Subsequently, the proposed method trains a Bayesian model, such as the Gaussian process regression~(GPR) model,
which predicts the objective values $y^{\rm{dataset}}$ according to the latent parameters ($z_{\rm{v}}^{\rm{dataset}}, z_{\rm{f}}^{\rm{dataset}}$).
The GPR model provides the predicted means $\mu(z_{\rm{v}}, z_{\rm{f}})$ and variances $\sigma(z_{\rm{v}}, z_{\rm{f}})$ of the objective values, which are used to compute the acquisition function $a(z_{\rm{v}}, z_{\rm{f}})$.
Bayesian optimization generates a latent variable parameter $z_{\rm{v}}^{\rm{target}}$ that maximizes the acquisition function under the equality constraint $z_{\rm{f}}=z_{\rm{f}}^{\rm{target}}$ in Equation~(\ref{equ:equalityConstraint}).
The  parameters $z_{\rm{v}}^{\rm{target}}$ are decoded into variable parameters $x_{\rm{v}}^{\rm{target}}$.

\begin{equation}
    \label{equ:equalityConstraint}
    z_{\rm{v}}^{\rm{target}} = \argmax a(\mu(z_{\rm{v}}, z_{\rm{f}}=z_{\rm{f}}^{\rm{target}}), \sigma(z_{\rm{v}}, z_{\rm{f}}=z_{\rm{f}}^{\rm{target}}))
\end{equation}

%% file: 4-dataset.tex
\section{Usage Scenario: Powder Weighing System}\label{sec:powderWeightingDevice}
\figpowder

\subsection{System Overview}\label{sec:system_overview}

As illustrated in Figure~\ref{fig:powder_weighing}, a powder weighing system weighs a user-specified amount of powder from the upper container into the lower container. 
The scale beneath the lower container weighed the amount of powder inside the lower container.
The system can control the amount of powder entering the lower container by vertically moving the valve of the upper container.
The degree to which the valve opens is referred to as the \emph{valve opening degree}.
This system adjusts the valve opening degree according to a preset parameter.
The preset parameter contains ten valve opening degrees $\{v_i\}$ and nine \emph{switching weights} $\{s_i\}$.
They configure nine pairs of them $\{(v_i, s_i)\}$ and an initial valve opening degree $v_0$.
The system starts by opening the valve according to the initial degree $v_0$.
When the weighing value reaches the switching weight $s_i$, it adjusts the valve to the corresponding degree $v_i$. 
Their proper values depend on the type of powder and the user-specified required weight.
It requires an average of 20 trials to obtain proper values, even for knowledgeable users.
A single trial took a maximum of five min.


The valve opening degrees and switching weights must satisfy two types of inequality constraints: non-negative and monotonically decreasing constraints.
The valve opening degrees generally have a larger value at the beginning of weighing and a smaller value at the end.
This is because the small valve opening degrees at the beginning takes a longer time to weigh the powder, 
and the large valve opening degrees at the end weigh more powder than expected.

\subsection{Dataset}\label{sec:dataset}
\tabparameter


This section reviews the various parameters included in the datasets for the powder weighing system and categorizes them into variable and fixed parameters, and objective values, as shown in Table~\ref{tab:parameter}.
The valve opening degrees and switching weights are variable parameters; that is, variable parameters have 19 dimensions.
The fixed parameters include 11 physical properties\footnote{Including the following properties: average particle size$[\mathrm{\mu m}]$, apparent specific gravity~(loose)[-], apparent specific gravity~(firm)[-], compressibility$[\%]$, angle of repose$[^\circ]$, spatula angle (average)$[^\circ]$, flowability index[-], collapse angle$[^\circ]$, difference angle$[^\circ]$, dispersion$[\%]$, jetting index[-]}, such as the jetting index that explains the ease of dispersibility of the powder.
The physical parameters are related to the fall speed of the powder.
For example, the larger the jetting index, the faster the fall speed of the powder is like a liquid because it easily contains air.
That is, a larger jetting index complicates the control of the fall speed of the powder.
The fixed parameters also include user-specified setting parameters for the powder weighing system, such as the required weight of the powder and whether to vibrate.
Three parameters are also related to the fall speed of the powder: shaking, vibration, and pre-vibration.
However, they are usually predetermined by knowledgeable users and rarely change; therefore, we set them as fixed parameters.
The absolute value of the difference between the measured and required weights, referred to as the weighing error, was defined as the objective value.


The datasets contained 60 types of powder and consisted of 1{,}792 trials.
The average number of trials for each powder was $31.33 \pm 19.48$.
The variable parameters that were finally adopted were also included in the dataset.
The criterion for parameter adoption was whether the ratio of the weighing error to the required weight was less than 1\%.

%% file: 5-implementation.tex
\section{Implementation}\label{sec:implementation}
\subsection{Preprocessing}\label{sec:preprocessing}

This section describes the preprocessing of the powder weighing system.
The following preprocessing steps are common to Steps 1 and 2, detailed in Section~\ref{sec:method}, except for duplication removal and data filtering.

\noindent
\textbf{Normalization}
Data normalization is an essential preprocessing step in machine learning.
Based on a pilot study, we normalized each parameter as follows.
For the fixed parameters, we performed a min-max normalization for each dimension.
For the variable parameters, we performed the same normalization for the first step of the valve opening degree $v_0$ and switching weight $s_1$ (Figure~\ref{fig:powder_weighing}).
However, we used the relative value of the first step in subsequent steps.
This is because the scale of the parameter values depends on the value of the first step.
Objective values were standardized using the mean and standard deviation.

\noindent
\textbf{Outlier removal}
We defined outliers as those that satisfied the following criteria: trials with large weighing errors or those whose valve opening degrees or switching weights did not decrease monotonically. 
Sixty-one trials were considered outliers.

\noindent
\textbf{Duplication removal}
Before Step 1, we removed trials whose fixed or variable parameters had the same values.
Because the same powder had similar fixed or variable parameters, the powder with fewer trials had more significant reconstruction errors than those with more trials.
After removing duplicates, we had 158 and 199 trials for the fixed and variable parameters, respectively.

\noindent
\textbf{Data filtering}
We need to define a bounding box as a search area in the latent space to perform Bayesian optimization in Step 2.
To optimize the variable parameters in a few trials, we assumed that powders with similar fixed parameters have similar optimal values for variable parameters.
Based on this assumption, we filtered the types of powder in the datasets whose fixed parameters were similar to those of the target powder $x_{\rm{f}}^{\rm{target}}$ and defined a bounding box that contained their latent variable parameters $z_{\rm{v}}^{\rm{dataset}}$.
This filtering improves the efficiency of parameter exploration because the search space is localized and the new trial has a more significant impact on GPR training.
We used the top seven powders whose fixed parameters were close to the target values based on Euclidean distance.
After filtering, data for approximately 200 trials were retained.

\noindent
\textbf{Train/Test split}
We divided trials in the datasets into training and test trials to avoid overfitting.
For Steps 1 and 2, 70 and 80\% of the trials in the datasets were used for training, respectively.

\subsection{Model Architecture and Hyperparameters}\label{sec:model}

This section describes the architecture and hyperparameters of the powder weighing system in Steps 1 and 2.


\noindent
\textbf{$\beta$-VAE}
For variable parameters, we designed the architecture of both the encoder and decoder in the $\beta$-VAE with three fully connected layers with 19 units and two ReLU activation layers.
We set the prior distribution in the latent space to be a standard normal distribution and the number of dimensions of the latent space to be two (detailed in Section~\ref{sec:ex1}).
We used the Adam~\cite{Kingma15} optimizer and set an initial learning rate to be 0.001.
Using these settings, we trained 1{,}000 epochs with a batch size of 32 and saved the model for the epoch when the loss was the lowest.
Model training completes in 1.5 min using a single GPU (NVIDIA GeForce RTX2070).


\noindent
\textbf{PCA}
We used PCA as a dimensionality reduction model for the fixed parameters.
Notably, the physical properties (Table~\ref{tab:parameter}) and the remaining parameters were encoded separately.
This is because the cumulative contribution ratio was approximately 60\% when compressing together into three dimensions, whereas it was approximately 98\% when compressing individually into one and two dimensions.
Therefore, the correlations between the physical properties and the others are weak.


\noindent
\textbf{GPR}
A kernel function of GPR is the weighted sum of the Matern 5/2 kernel and automatic relevance determination, and the weight is estimated based on the maximum likelihood.


\noindent
\textbf{Acquisition function}
We used the upper confidence bound~(UCB)~\cite{Peter03} as the acquisition function to interactively allow users to change an optimization strategy.
Users can choose one of three types of variable parameters: exploration-oriented, exploitation-oriented, or intermediate.
The corresponding UCB hyperparameters were set to 0.001, 1.0, and 0.5, respectively, based on a melody design application~\cite{Zhou21}.

%% file: 6-evaluation.tex
\section{Evaluation}\label{sec:evaluation}


\subsection{Experiment1: Effectiveness of Step 1}\label{sec:ex1}
\subsubsection{Procedure}

The aim of Experiment 1 is to verify whether the introduction of DRL enables us to consider inequality constraints.
We randomly sampled many data samples from the latent space and counted the number of samples that did not satisfy the inequality constraints.
We evaluated the correlations between the dimensionality of the latent space $d_{\rm{v}}$, the value of the coefficient $\beta$, and the number of samples that did not satisfy the inequality constraints.
The inequality constraints include non-negative and monotonically decreasing constraints, which users do not explicitly specify to train the dimensionality reduction models.
We also visualized the variable parameters of the samples and confirmed the features acquired by DRL for a qualitative evaluation.
Based on the results, we discuss the proper dimensionality of $\beta$-VAE and the appropriate coefficient $\beta$ for Experiment 2.

\figexratio
\figexparameter
\subsubsection{Result}


Figure~\ref{fig:ex1_ratio} shows the relationship between coefficient $\beta$ and the number of samples that do not satisfy the inequality constraints ((a)~non-negative constraint, (b)~monotonically decreasing constraint, and (c) (a) or (b)).
We sampled 1{,}000 data samples within the hypersphere with a radius of two around the origin of the latent space.
The experimental results are as follows:
\begin{itemize}
    \item A larger coefficient $\beta$ decreases the ratio of data samples that do not satisfy the inequality constraints, implying that the introduction of DRL enables us to consider the inequality constraints.
    \item Larger dimensionality in the latent space increases the ratio of data samples that do not satisfy the inequality constraints.
    Based on our observations, samples far from the origin of the latent space do not satisfy the inequality constraints because the large dimensionality learns undesirable fine-grained features and is emphasized in the area far from the origin.
    \item Because (b) and (c) show almost the same trend, data samples that do not satisfy the non-negative constraint do not meet the monotonically decreasing constraint. 
\end{itemize}


Next, to determine the proper coefficient $\beta$ of $\beta$-VAE, we compared the valve opening degrees and switching weights reconstructed from the two-dimensional latent space in terms of $\beta=0.1, 0.5, 1.0$.
Figure~\ref{fig:ex1_parameter} shows 15 types of variable parameters sampled in the range of $[-2, 2]$ along each axis of the two-dimensional latent space.
They imply that a larger $\beta$ eases the satisfaction of the inequality constraints, but produces similar variable parameters, called posterior collapse.
Posterior collapse increases reconstruction errors and we may be unable to determine the proper parameters even after many trials.
Therefore, posterior collapse is more severe than when inequality constraints are not satisfied.
Consequently, we decided to use $\beta=0.1$ in Experiment 2.

\subsection{Experiment2: Effectiveness of Step 2}\label{sec:ex2}
\subsubsection{Procedure}
\tabcondition
\figexsecond

Experiment 2 aims to verify whether the proposed method could determine optimum parameters within a reasonable number of trials.
The number of trials required to achieve the target weighing error was less than 1\% of the required weight of the powder.
In this experiment, we used three types of powders not included in the datasets: Powders A, B, and C.
Their features are as follows:
\begin{itemize}
    \item The powders are highly floodable.
    \item Powders A and B are more fluid than powder C.
    \item The grain diameter is larger in the order of powders A, B, and C.
\end{itemize}
Table~\ref{tab:condition} lists the user-specified fixed parameters of each powder.
Figure~\ref{fig:embedding} maps 60 types of powder in the datasets on a 2D scatterplot according to their fixed parameters and indicates that powders A, B, and C can appropriately validate the versatility of the proposed method because none of them are outliers.


The parameters were explored based on the users' experience and expertise to evaluate the system in an actual operation.
The following steps were performed by users for the exploration:
\begin{itemize}
    \item Users choose one of the three candidates as a variable parameter: exploitation-oriented, exploration-oriented, and intermediate, as detailed in Section~\ref{sec:model}.
    \item When parameters that do not satisfy the inequality constraints are generated, a minimum modification is made to the parameters to satisfy the inequality constraints before evaluation. 
    When parameters require extensive modification or are invalid, users penalize them with a 10\% weighing error without trying them.
\end{itemize}

\subsubsection{Result}

Figure~\ref{fig:bo_result} shows the ratio of the weighing errors to the required weight of the powder through seven trials for the three powders.
We achieved the target weighing error (less than 1\% of the required weight) within seven trials, which is approximately one-third of the number of trials required for manual tuning.
Based on the results, the proposed method worked effectively and $\beta$-VAE and GPR models have a high generalization performance for unknown powders.


We successfully generated variable parameters that satisfy the two inequality constraints in all trials for powders B and C.
In contrast, for Powder A, the proposed method generated variable parameters that did not satisfy the non-negative constraint in one trial and were penalized without evaluation in two trials. 
The proposed method seems to have explored areas far from the origin of the latent space, as shown in Figure~\ref{fig:embedding}.

%% file: 7-discussion.tex
\section{Discussion}\label{sec:discussion}
\subsection{Hyperparameter Tuning}\label{sec:hyperparameter_tuning}

As illustrated in Experiment 1, two hyperparameters of $\beta$-VAE should be adjusted: the number of dimensions of the latent space $d_{\rm{v}}$ and coefficient $\beta$.
A larger $d_{\rm{v}}$ leads to a small reconstruction error, whereas it generates variable parameters that are unlikely to satisfy the inequality constraints.
However, when coefficient $\beta$ is large, a posterior collapse that increases the reconstruction error occurs and variable parameters are more likely to satisfy the inequality constraints.
These contrasting effects indicate that the appropriate value of coefficient $\beta$ depends on the dimensions $d_{\rm{v}}$.
Therefore, the appropriate values of them can be determined using Figures~\ref{fig:ex1_ratio} and \ref{fig:ex1_parameter} because they visualize a trade-off between the effects.
The results of Experiment 2 validate that appropriate hyperparameter tuning contributes to achieving the target performance in limited trials.


However, the robustness of the setting of the two hyperparameters remains unclear because Experiment 2 was only tested for a specific combination.
Other combinations of hyperparameters will be used in the experiment and methods to improve the robustness will be investigated.
Particularly, we will consider applying other DRL methods, such as FactorVAE~\cite{Kim18} and $\beta$--TCVAE~\cite{Chen18}, which improve the trade-off between the reconstruction error and the Kullback-Liebrier divergence.

\subsection{Search Area in Latent Space}\label{sec:search_area}

Bayesian optimization requires defining a search area in a latent space.
However, it is not trivial to determine the search area.
As described in Section~\ref{sec:implementation}, we assumed that different types of powders with similar fixed parameters have similar optimal values for variable parameters and only the powders with similar fixed parameters are filtered.
Therefore, the search area is localized compared with the case without filtering.
This assumption is a slightly strong restriction because the optimal parameters may not be in the search area.
The search range should be large according to the user-acceptable number of trials.
However, suppose the search area is too large.
In this case, the proposed method may generate variable parameters that do not satisfy the inequality constraints because of the emphasis on fine-grained features, as shown in Figure~\ref{fig:ex1_ratio} (a).
To address this challenge, we investigated methods to automatically determine the search area.
We believe that a method~\cite{Pascal21} that narrows the search area based on decoder uncertainty should be used.

\subsection{Generalizability}\label{sec:future_work}

From this empirical study, we will extend the generalizability of the proposed method from the perspective of constraints.
The proposed method cannot consider unknown equality constraints, known inequality constraints, or more complicated constraints, such as the case in which inequality constraints depend on fixed parameters.
We will also apply the proposed method to other tasks, such as molecule design~\cite{Bombarelli18}\cite{Yan20}, which handles discrete parameters.

%% file: 8-conclusion.tex
\section{Conclusion}\label{sec:conclusion}

This study designed a general framework that combines dimensionality reduction and Bayesian optimization for high-dimensional parameters with constraints.
As observed, introducing DRL into dimensionality reduction enables us to consider inequality constraints, even if users do not explicitly provide them.
We applied the proposed method to a powder weighing system and proved that the method could implicitly consider inequality constraints, including non-negative and monotonically decreasing constraints.
Moreover, the proposed method successfully optimized parameters within a reasonable number of trials and reduced the required number of trials by 66\% compared with manual parameter tuning.
Experiments also validated the importance of two hyperparameters: dimensionality in the latent space and coefficient $\beta$ of $\beta$-VAE.
The robustness of the proposed method for these hyperparameters and a method to automatically determine the search area for Bayesian optimization will be examined in future studies.

%% file: author.bbl
\begin{thebibliography}{10}
\providecommand{\url}[1]{{#1}}
\providecommand{\urlprefix}{URL }
\expandafter\ifx\csname urlstyle\endcsname\relax
  \providecommand{\doi}[1]{DOI~\discretionary{}{}{}#1}\else
  \providecommand{\doi}{DOI~\discretionary{}{}{}\begingroup
  \urlstyle{rm}\Url}\fi

\bibitem{Akiba19}
Akiba, T., Sano, S., Yanase, T., Ohta, T., Koyama, M.: Optuna: A
  next-generation hyperparameter optimization framework.
\newblock In: KDD '19, p. 2623–2631

\bibitem{Peter03}
Auer, P.: Using confidence bounds for exploitation-exploration trade-offs
  \textbf{3}, 397–422 (2003)

\bibitem{Balandat20}
Balandat, M., Karrer, B., Jiang, D.R., Daulton, S., Letham, B., Wilson, A.G.,
  Bakshy, E.: {BoTorch: A Framework for Efficient Monte-Carlo Bayesian
  Optimization}.
\newblock In: NeurIPS '20

\bibitem{Bergmann20}
Bergmann, D., Graichen, K.: Safe bayesian optimization under unknown
  constraints.
\newblock In: CDC '20, pp. 3592--3597

\bibitem{Chen18}
Chen, R.T.Q., Li, X., Grosse, R., Duvenaud, D.: Isolating sources of
  disentanglement in variational autoencoders.
\newblock In: NeurIPS '18

\bibitem{Bombarelli18}
Gómez-Bombarelli, R., Wei, J.N., Duvenaud, D., Hernández-Lobato, J.M.,
  Sánchez-Lengeling, B., Sheberla, D., Aguilera-Iparraguirre, J., Hirzel,
  T.D., Adams, R.P., Aspuru-Guzik, A.: Automatic chemical design using a
  data-driven continuous representation of molecules.
\newblock ACS Central Science \textbf{4}(2), 268--276 (2018)

\bibitem{Hansen20}
Hansen, N., Auger, A., Ros, R., Mersmann, O., Tu{\v{s}}ar, T., Brockhoff, D.:
  {COCO}: a platform for comparing continuous optimizers in a black-box
  setting.
\newblock Optimization Methods and Software \textbf{36}(1), 114--144 (2020)

\bibitem{Higgins17}
Higgins, I., Matthey, L., Pal, A., Burgess, C.P., Glorot, X., Botvinick, M.M.,
  Mohamed, S., Lerchner, A.: beta-vae: Learning basic visual concepts with a
  constrained variational framework.
\newblock In: ICLR '17

\bibitem{Jaquier20}
Jaquier, N., Rozo, L.: High-dimensional bayesian optimization via nested
  riemannian manifolds.
\newblock In: NeurIPS '20

\bibitem{Jolliffe86}
Jolliffe, I.: Principal Component Analysis.
\newblock Springer Verlag (1986)

\bibitem{Kandasamy15}
Kandasamy, K., Schneider, J., Poczos, B.: High dimensional bayesian
  optimisation and bandits via additive models.
\newblock In: ICML '15, vol.~37, pp. 295--304

\bibitem{Kim18}
Kim, H., Mnih, A.: Disentangling by factorising.
\newblock In: ICML '18, vol.~80, pp. 2649--2658

\bibitem{Kingma15}
Kingma, D.P., Ba, J.: Adam: {A} method for stochastic optimization.
\newblock In: ICLR '15

\bibitem{Kingma14}
Kingma, D.P., Welling, M.: {Auto-Encoding Variational Bayes}.
\newblock In: ICLR '14

\bibitem{Kirschner19}
Kirschner, J., Mutny, M., Hiller, N., Ischebeck, R., Krause, A.: Adaptive and
  safe {B}ayesian optimization in high dimensions via one-dimensional
  subspaces.
\newblock In: ICML '19, vol.~97, pp. 3429--3438

\bibitem{Letham20}
Letham, B., Calandra, R., Rai, A., Bakshy, E.: Re-examining linear embeddings
  for high-dimensional {B}ayesian optimization.
\newblock In: NeurIPS '20

\bibitem{Li17}
Li, C., Gupta, S., Rana, S., Nguyen, V., Venkatesh, S., Shilton, A.: High
  dimensional bayesian optimization using dropout.
\newblock In: IJCAI '17, pp. 2096--2102

\bibitem{Li16}
Li, C.L., Kandasamy, K., Poczos, B., Schneider, J.: High dimensional bayesian
  optimization via restricted projection pursuit models.
\newblock In: AISTAT '16

\bibitem{Liaw18}
Liaw, R., Liang, E., Nishihara, R., Moritz, P., Gonzalez, J.E., Stoica, I.:
  Tune: A research platform for distributed model selection and training
  (2018)

\bibitem{Lu18}
Lu, X., Gonzalez, J., Dai, Z., Lawrence, N.: Structured variationally
  auto-encoded optimization.
\newblock In: ICML '18, vol.~80, pp. 3267--3275

\bibitem{Moriconi20}
Moriconi, R., Deisenroth, M., K~S, S.K.: High-dimensional bayesian optimization
  using low-dimensional feature spaces.
\newblock Machine Learning \textbf{109}, 1925--1943 (2020)

\bibitem{Pascal21}
Notin, P., Hern\'{a}ndez-Lobato, J.M., Gal, Y.: Improving black-box
  optimization in vae latent space using decoder uncertainty.
\newblock In: NeurIPS '21, vol.~34, pp. 802--814

\bibitem{Rolland18}
Rolland, P., Scarlett, J., Bogunovic, I., Cevher, V.: High-dimensional bayesian
  optimization via additive models with overlapping groups.
\newblock In: AISTATS '18

\bibitem{Wang17}
Wang, Z., Li, C., Jegelka, S., Kohli, P.: Batched high-dimensional bayesian
  optimization via structural kernel learning.
\newblock In: ICML'17, p. 3656–3664

\bibitem{Wang13}
Wang, Z., Zoghi, M., Hutter, F., Matheson, D., De~Freitas, N.: Bayesian
  optimization in high dimensions via random embeddings.
\newblock In: IJCAI '13, p. 1778–1784

\bibitem{Yan20}
Yan, C., Wang, S., Yang, J., Xu, T., Huang, J.: Re-balancing variational
  autoencoder loss for molecule sequence generation.
\newblock In: BCB '20

\bibitem{Zhou21}
Zhou, Y., Koyama, Y., Goto, M., Igarashi, T.: Interactive
  exploration-exploitation balancing for generative melody composition.
\newblock In: IUI '21, p. 43–47

\end{thebibliography}
